\documentclass{esannV2}
\usepackage[dvips]{graphicx}
\usepackage[latin1]{inputenc}
\usepackage{amssymb,amsmath,array}
\usepackage{amsthm}
\usepackage{caption}

\usepackage{footnote}
\newcommand{\refdef}[1]{Definition~\ref{#1}}
\newcommand{\reflemma}[1]{Lemma~\ref{#1}}

\newcommand{\refeq}[1]{Eq.~\eqref{#1}}

\usepackage{listings}
\usepackage{ifxetex}
\usepackage{fancyhdr}
\usepackage{hyperref}
\usepackage{graphicx}
\usepackage{wrapfig}
\usepackage{amsmath}
\usepackage{amsfonts}
\usepackage{amssymb}
\usepackage{listings}
\usepackage{setspace}
\usepackage{wrapfig}
\usepackage{tablefootnote}
\usepackage{multirow}
\usepackage{booktabs}
\usepackage[section]{placeins}
\usepackage{color}
\usepackage{amsfonts}
\usepackage{ulem}
\usepackage{algorithm}
\usepackage{algorithmicx}
\usepackage{algpseudocode}
\usepackage{graphicx}
\usepackage{float}
\usepackage{dsfont}
\usepackage{cleveref}

\newcommand{\mat}[1]{\mathbf{#1}}
\newcommand{\set}[1]{\mathcal{#1}}
\newcommand{\pnorm}[1]{\lVert{#1}\rVert}

\DeclareMathOperator*{\loss}{{\ell}}

\DeclareMathOperator*{\trace}{{trace}}

\DeclareMathOperator*{\SetSymMat}{\mathcal{S}}

\DeclareMathOperator*{\diag}{diag}

\newcommand{\E}{\mathbb{E}}
\newcommand{\RN}{\mathbb{R}}

\DeclareMathOperator*{\densitythreshold}{\ensuremath{\delta}}
\newcommand{\density}{\ensuremath{p}}

\DeclareMathOperator*{\regularization}{\ensuremath{{\theta}}}
\newcommand{\x}{\ensuremath{\vec{x}}}
\newcommand{\X}{\ensuremath{\vec{X}}}

\newcommand{\xorig}{\ensuremath{\vec{x}_{\text{orig}}}}

\newcommand{\sety}{\ensuremath{\set{Y}}}

\newcommand{\xcf}{\ensuremath{\vec{x}_{\text{cf}}}}

\newcommand{\ycf}{\ensuremath{y_{\text{cf}}}}
\newcommand{\deltacf}{\ensuremath{\vec{\delta}}}

\newcommand{\dimsym}{d}
\newcommand{\latenspacedim}{m}
\newcommand{\classifier}{\ensuremath{h}}

\newcommand{\CovMat}{\ensuremath{\mat{\Sigma}}}
\newcommand{\CovMatEmp}{\ensuremath{\mat{\Sigma}_{\text{emp}}}}
\newcommand{\CorrMat}{\ensuremath{\mat{\tilde{\Sigma}}}}
\newcommand{\actionfunc}{\ensuremath{f}}
\newcommand{\ProtoMat}{\ensuremath{\mat{P}}}
\newcommand{\ProtoBase}{\ensuremath{\vec{b}}}
\newcommand{\encodingfunc}{\ensuremath{z}}
\newcommand{\covx}{\ensuremath{\text{covx}}}

\usepackage{amsthm}
\newtheorem{theorem}{Theorem}
\newtheorem{definition}{Definition}
\newtheorem{lemma}[theorem]{Lemma}

\begin{document}
\theoremstyle{remark}
\newtheorem{remark}{Remark}

\title{Convex optimization for actionable \& plausible counterfactual explanations}

\author{Andr\'e Artelt\footnote{Corresponding author: \href{mailto:aartelt@techfak.uni-bielefeld.de}{aartelt@techfak.uni-bielefeld.de}}\; and Barbara Hammer
%
\thanks{We gratefully acknowledge funding from the VW-Foundation for the project \textit{IMPACT} funded in the frame of the funding line \textit{AI and its Implications for Future Society}.}
%
\vspace{.3cm}\\
%
CITEC - Cognitive Interaction Technology \\
Bielefeld University - Faculty of Technology \\
Inspiration 1, 33619 Bielefeld - Germany
%
}

\maketitle

\begin{abstract}
Transparency is an essential requirement of machine learning based decision making systems that are deployed in real world. Often, transparency of a given system is achieved by providing explanations of the behaviour and predictions of the given system.
Counterfactual explanations are a prominent instance of particular intuitive explanations of decision making systems.
While a lot of different methods for computing counterfactual explanations exist, only very few work (apart from work from the causality domain) considers feature dependencies as well as plausibility which might limit the set of possible counterfactual explanations.

In this work we enhance our previous work on convex modeling for computing counterfactual explanations by a mechanism for ensuring actionability and plausibility of the resulting counterfactual explanations.
\end{abstract}

\section{Introduction}\label{sec:introduction}
Nowadays we are faced with an increasing deployment of machine learning (ML) and artificial intelligence (AI) based decision making systems in the real world - e.g. predictive policing~\cite{PredictivePolicing} and loan approval~\cite{CreditRiskML,CreditScoresUnfair}. Because of the high impact of many of these systems, policy makers demand transparency and interpretability of such decision making systems - first approaches have already been manifested in legal regulations like the EUs GDPR~\cite{gdpr}.
There exist a wide variety of methods for explaining ML and AI based decision making systems and thus meeting the demands for transparency and interpretability. A popular class of explanation methods, that are not tailored to a specific model but rather universal, are model agnostic methods: Feature interaction methods~\cite{featureinteraction}, feature importance methods~\cite{featureimportance} and example based methods~\cite{casebasedreasoning}.
Popular instances of example based methods are influential instances~\cite{influentialinstances}, prototypes \& criticisms~\cite{prototypescriticism} and counterfactual explanations~\cite{counterfactualwachter}.
A counterfactual explanation is a change of the original input that leads to a different (specific) prediction or behavior of the decision making system - \textit{what has to be different in order to change the prediction of the system?}
Such an explanation is considered to be intuitive and useful because it proposes changes to achieve a desired outcome, i.e. it provides actionable feedback~\cite{molnar2019,counterfactualwachter}. Furthermore, there exists strong evidence that explanations by humans are often counterfactual in nature~\cite{ijcai2019-876}. We will focus on theses types of explanations in this work.

\section{Counterfactual explanations}\label{sec:counterfactuals}
Counterfactual explanations (often just called counterfactuals) contrast samples by counterparts with minimum change of the appearance but different class label~\cite{counterfactualwachter,molnar2019} and can be formalized as follows: 
\begin{definition}[Counterfactual explanation~\cite{counterfactualwachter}]\label{def:counterfactual}
Let $\classifier:\RN^\dimsym \to \sety$ be a given prediction function. Computing a counterfactual $\xcf \in \RN^\dimsym$ for a given input $\xorig \in \RN^\dimsym$ is phrased as an optimization problem:
\begin{equation}\label{eq:counterfactualoptproblem}
\underset{\xcf \,\in\, \RN^\dimsym}{\arg\min}\; \loss\big(\classifier(\xcf), \ycf\big) + C \cdot \regularization(\xcf, \xorig)
\end{equation}
where $\loss(\cdot)$ denotes a suitable loss function, $\ycf$ the requested prediction, and  $\regularization(\cdot)$  a penalty term for deviations of $\xcf$ from the original input $\xorig$. $C>0$ denotes the regularization strength.
\end{definition}
While the classical formalization~\refeq{eq:counterfactualoptproblem} is model agnostic - i.e. it does not make any assumptions on the model $\classifier(\cdot)$ -, it can be beneficial to rewrite the optimization problem~\refeq{eq:counterfactualoptproblem} in constrained form~\cite{counterfactualcomputationsurvey}:
\begin{subequations}\label{eq:cf:constraintform}
\begin{align}
& \underset{\xcf \,\in\,\RN^\dimsym}{\arg\min}\;\regularization(\xcf, \xorig) \label{eq:cf:constraintform:objective} \\
& \text{ s.t. } \classifier(\xcf) = \ycf  \label{eq:cf:constraintform:constraint}
\end{align}
\end{subequations}
In our previous work~\cite{counterfactualcomputationsurvey} we have shown that the constrained optimization problem~\refeq{eq:cf:constraintform} can be turned (or efficiently approximated) into convex programs for many standard machine learning models including generalized linear models, quadratic discriminant analysis, nearest neighbor classifiers, etc. Since convex programs can be solved efficiently~\cite{Boyd2004}, the constrained form~\cite{counterfactualcomputationsurvey} becomes superior over the original black-box modeling~\cite{counterfactualwachter} if we have access to the underlying model $\classifier(\cdot)$.

The counterfactuals from~\refdef{def:counterfactual} are also called closest counterfactuals because they try to find a point that is as close as possible to the original sample. Is was observed that this often leads to adversarials which might not be useful for explaining the models behaviour~\cite{plausiblecounterfactualsartelt}. When this is an issue, additional plausibility constraints are added~\cite{counterfactualcomputationsurvey,chou2021counterfactuals,face,plausiblecounterfactualsartelt} often modeled based on densities like stated in~\refdef{def:plausiblecounterfactual}.
\begin{definition}[$\densitythreshold$-plausible counterfactual~\cite{plausiblecounterfactualsartelt}]\label{def:plausiblecounterfactual}
Let $\classifier:\RN^\dimsym  \to \sety$ be a prediction function and $\density(\cdot)$ a class dependent density. We call a counterfactual explanation $(\xcf, \ycf)$ of a particular sample $\x\in\RN^\dimsym$ $\densitythreshold$-plausible iff the following holds:
\begin{equation}\label{eq:deltaplausiblecounterfactual:optprob}
\xcf = \underset{\xcf \,\in\,\RN^\dimsym}{\arg\min}\;\regularization(\xcf, \x) \quad \text{s.t. } \classifier(\xcf) = \ycf \; \text{ and } \; \density(\xcf;\ycf) \geq \densitythreshold
\end{equation}
\end{definition}
where $\densitythreshold > 0$ denotes a minimum density at which we consider a sample plausible.

When it comes to communicate the explanation to the use, we have two possibilities:
We can either present the solution $\xcf$ to the user or the difference $\deltacf = \xorig - \xcf$ as an explanation to the user. Since the popularity of counterfactual explanations comes from the recommendation of actions that lead to a desired goal (actionable feedback to the user), it is natural to communicate the difference $\deltacf$ to the user. However, when computing and communicating the difference $\deltacf$, we implicitly assume that all features are independent of each other - i.e. we assume that we can arbitrarily and independently change all features at the same time. This is a strong assumption which might not always be true in practice - e.g. it could be the case that some features are anti-correlated so that they can not be both increased or decreased at the same time.
Another problem with the plain difference $\deltacf$ as an explanation is that it assumes that the features are somewhat meaningful and interpretable, which is not always the case like in case of images\footnote{In case of images, we have pixel as features which are not informative at all.}.

\paragraph*{Actionable and plausible counterfactuals}
Instead of directly optimizing the final counterfactual $\xcf$, we propose to optimize over an \textit{action vector} which encodes the actions/changes applied to the original sample $\xorig$ which then yield the final counterfactual $\xcf$. We define a function that maps a given action vector to a (potentially counterfactual) explanation $\xcf$ - note that $\xcf$ might not necessarily be a valid counterfactual because the property of being a counterfactual depends on a prediction function $\classifier(\cdot)$ which the function $\actionfunc(\cdot)$ is not aware of, $\actionfunc(\cdot)$ only applies an action vector to the original sample $\xorig$. 
\begin{equation}\label{eq:actionfunction}
    \actionfunc:\deltacf \mapsto \xcf
\end{equation}
Since~\refeq{eq:actionfunction} should depend on the original sample $\xorig$, we explicitly note the dependency on $\xorig$ by writing:
\begin{equation}
    \actionfunc(\deltacf;\xorig)=\xcf
\end{equation}
The action vector $\deltacf$ can live in the same space like the original sample $\xorig$ - i.e. $\deltacf$ and $\xorig$ share the same features. However, it is also possible that $\deltacf$ lives in some kind of latent space - i.e. changing some abstract attributes.
In other words, the function $f(\cdot)$ applies the changes $\deltacf$ to the original sample $\xorig$ and embeds the result in the data space that is used for making predictions - i.e. it might have to embed text into some vector space, consider feature dependencies, apply one-hot-encodings, etc.

The final optimization problem for computing an action vector that leads to a valid counterfactual explanation is phrased as the following optimization problem:
\begin{subequations}\label{eq:counterfactualopt:cfactionable}
\begin{align}
&\underset{\deltacf \,\in\,\RN^\latenspacedim}{\arg\min}\;\regularization(\deltacf) \label{eq:cfactionable:constraintform:objective} \\
& \text{ s.t. } \classifier\left(\actionfunc(\deltacf;\xorig)\right) = \ycf  \label{eq:cfactionable:constraintform:constraint}
\end{align}
\end{subequations}
The original modeling~\refeq{eq:cf:constraintform} can be obtained as a special case of ~\refeq{eq:counterfactualopt:cfactionable} if $\actionfunc(\deltacf;\xorig)=\xorig + \deltacf$ and $\regularization(\cdot)$ denotes a suitable metric.

In this work, we focus on two aspects: Feature dependencies and plausibility.
In section~\ref{sec:modeling:featuredependecies} we propose a realization of $\actionfunc(\cdot)$ that takes care of potential feature dependencies - i.e. feature can not be changed independently.
And in section~\ref{sec:modeling:plausibility} we consider a realization of $\actionfunc(\cdot)$ where the action vectors corresponds to selecting primitives/prototypes that are combined into the final sample - i.e. some kind of sparse coding via a given codebook (similar to a latent space approach).
Finally, in section~\ref{sec:experiments} we empirically evaluate our proposed modelings on several data sets and models.

\paragraph*{Related work}
As already mentioned, there exist a wide variety of work that deals with the computation of counterfactual explanations~\refdef{def:counterfactual} as well as considering additional aspects like plausibility~\cite{counterfactualcomputationsurvey,chou2021counterfactuals,face,plausiblecounterfactualsartelt}. However, usually these methods assume that all features can be changed independently from each other - i.e. no further constraints are posed on the actionability of the difference $\deltacf= \xorig - \xcf$ - only methods that use structural causal models or some predefined set of valid actions take potential feature dependencies into account~\cite{galhotra2021explaining,DBLP:journals/corr/abs-1912-03277}. In addition, those existing methods often have high computational complexity by relying on non-convex optimizations like integer programming.

A method for computing actionable recourse based on probabilistic counterfactual explanations is proposed in~\cite{galhotra2021explaining}. This method heavily relies on a given or estimated probabilistic causal model that encodes all variable dependencies - the counterfactuals itself are then computed by solving integer programs.

Another (early) work~\cite{DBLP:conf/fat/UstunSL19} is concerned with actionable recourse of linear classifiers. For a given set of actionable actions, their method solves an inter program for computing the final actionable counterfactual explanation.

In~\cite{face} a method called FACE for computing feasible and actionable counterfactual explanations is proposed. Instead of computing a single change or action vector that leads to the counterfactual, a path of intermediate samples is constructed that finally lead to the counterfactual explanation - assuming that moving from one sample to a nearby sample is always feasible and actionable.

The authors of~\cite{DBLP:journals/corr/abs-1912-03277} propose to use a variational autoencoder for learning feasible/actionability constraints from labeled data or user feedback. These learned constraints are then used when computing the final counterfactual explanations.

Other work like~\cite{DBLP:conf/ecai/FeghahatiSPT20,DBLP:journals/corr/abs-2103-10226,DBLP:journals/corr/abs-2012-09301} use some kind of autoencoder to learn a latent space and then compute the counterfactuals in this latent space to get plausible and meaningful counterfactual explanations.

All these methods need some kind causal graph as an input and are computational infeasible for many models or are even limited to linear models - on the other hand, methods that work for arbitrary models come without any formal guarantees.

%

\section{Actionable and plausible counterfactual explanations}\label{sec:generalidea}
\subsection{Feature dependencies}\label{sec:modeling:featuredependecies}
Instead of assuming independence of all features, we aim for a mechanism that allows us to consider potential feature dependencies when computing counterfactual explanations.

Assuming a fixed but unknown data generating process, the simplest measurement of feature dependencies is to consider the covariance matrix $\CovMat\in\SetSymMat_{+}^{\dimsym}$:
\begin{equation}\label{eq:covariancematrix}
    \CovMat = \E\Big[\big(\X - \E[\X]\big)\big(\X - \E[\X]\big)^\top\Big]
\end{equation}
where $\X$ denotes the random vector of the underlying generating process.

By standardizing the covariance~\refeq{eq:covariancematrix}, we obtain the correlation matrix $\CorrMat\in\SetSymMat_{+}^{\dimsym}$ which can be stated as follows:
\begin{equation}\label{eq:correlationmatrix}
    \CorrMat = \diag(\CovMat)^{-\frac{1}{2}}\CovMat\diag(\CovMat)^{-\frac{1}{2}}
\end{equation}
In the remainder of this work, we use the correlation matrix~\refeq{eq:correlationmatrix} for encoding feature dependencies - however, note that correlation does not necessarily denote a causal relationship - furthermore, non-linear dependencies are also not covered. We still think that considering only linear feature dependencies can already be beneficial and is better than assuming independence of all features which might be unrealistic for many real world scenarios.

Usually, the true covariance matrix is not known in practice and therefore we have to work with some kind of approximation. It might be the case that some entries in the covariance matrix are known or at least some educated guesses by domain experts are available. Otherwise we have to estimate it from a given data set. Since the straight forward estimation by using the maximum likelihood estimator\footnote{Also called empirical covariance where we replace the expectations by averages.} of~\refeq{eq:covariancematrix} is unstable in high dimensions, we use a method for estimating a sparse covariance matrix~\cite{sparsecovmat}\footnote{The choice of this particular method is kind of arbitrary - there exist many other methods for estimating covariance matricies in (possibly) high dimensional spaces.}:
\begin{equation}\label{eq:sparsecovmatestimator}
    \hat{\CovMat}^{-1} = \underset{\hat{\CovMat}^{-1} \,\in\,\SetSymMat_{+}^{\dimsym}}{\arg\min}\,\trace\left(\CovMatEmp \hat{\CovMat}^{-1}\right) - \log\left(\det(\hat{\CovMat}^{-1})\right) + \alpha\pnorm{\hat{\CovMat}^{-1}}_1
\end{equation}
where $\CovMatEmp\in\SetSymMat_{+}^{\dimsym}$ denotes the empirical covariance matrix that has been estimated from a given data set using the maximum likelihood estimator,$\pnorm{\cdot}_1$ denotes the sum of the absolute values of the off-diagonal coefficients of the given matrix and $\alpha>0$ is a hyperparameter that controls the sparsity of the covariance matrix (higher values lead to more sparsity).

\subsubsection{Realization of the action mapping function}
In order to ensure that the final counterfactual is actionable according to linear feature dependencies, we define the action mapping function $\actionfunc(\cdot)$ as follows: 
\begin{equation}\label{eq:cfparamaffinemapping}
    \actionfunc(\deltacf;\xorig)=\xorig + \CorrMat(\xorig)\deltacf
\end{equation}
where $\CorrMat(\xorig)$ denotes the correlation matrix which encodes the specific linear feature dependencies for the particular sample $\xorig$.
While one could use a global correlation matrix $\CorrMat$ which does not depend on $\xorig$, it might be beneficial to be able to use different correlation matrices for different users or groups. By this we can take differences in feasible actions for different users into account.

In the context of~\refeq{eq:cfparamaffinemapping}, changing some feature in $\deltacf$ might result in a change (increase or decrease) of a different feature - i.e. it might be impossible to increase two features at the same because of some anti-correlation. While this formalization~\refeq{eq:cfparamaffinemapping} allows us to encode some feature dependencies, note that it is still limited in the sense that time is ignored - i.e. all changes might not take place at the time but be applied in some order which itself could have some consequences on the previous changes.

\subsection{Plausibility}\label{sec:modeling:plausibility}
We ensure plausibility by assuming that all data samples are must be composed from a set of given primitives/prototypes arranged column wise in a matrix $\ProtoMat\in\RN^{\dimsym\times\latenspacedim}$ as well as a base $\ProtoBase\in\RN^\dimsym$. We therefore define $\actionfunc(\cdot)$ as follows:
\begin{equation}\label{eq:actionfunction:plausibility}
    \actionfunc(\deltacf;\xorig) = \ProtoMat\left(\encodingfunc(\xorig) + \deltacf\right) + \ProtoBase
\end{equation}
where $\encodingfunc: \RN^\dimsym \to \RN_+^\latenspacedim$ denotes a function that encodes the given sample in a latent space - i.e. the selection and weighting of the given primitives in $\ProtoMat$.
The primitives could be given (i.e. hand engineered) or learned by some kind of dictionary/representation learning (including PCA and ICA).

If it happens to be the case that all samples in the convex hull of the primitives $\covx(\ProtoMat,\ProtoBase)$ are $\delta$-plausible according to~\cite{plausiblecounterfactualsartelt}, we can easily ensure $\delta$-plausibility (under~\refdef{def:plausiblecounterfactual}) of the final counterfactual explanations as stated in~\reflemma{lemma:convexplausibleprimitives}.
\begin{lemma}\label{lemma:convexplausibleprimitives}
    Assuming that for given primitives $\ProtoMat,\ProtoBase$, all sample $\x \in\covx(\ProtoMat, \ProtoBase)$ are $\delta$-plausible (\refdef{def:plausiblecounterfactual}) for a fixed and given $\delta>0$.
    
    Then, when using~\refeq{eq:actionfunction:plausibility} as a realization of the action function~\refeq{eq:actionfunction} and the additional linear constraints $\sum_i(\encodingfunc(\xorig) + \deltacf)_i = 1$, $\deltacf\geq\vec{0}$, the resulting counterfactual explanations $\xcf$~\refeq{eq:counterfactualopt:cfactionable} are guranteed to be $\delta$-plausible.
\end{lemma}

\subsection{Convex optimization for actionable \& plausible counterfactuals}
In our previous work~\cite{counterfactualcomputationsurvey}, we show how to rewrite the constraint~\refeq{eq:cf:constraintform:constraint} as a set of convex constraints or a set of convex programs for many different ML models\footnote{For some models we propose model specific convex approximations.}.
We assume that we can rewrite the constraint~\refeq{eq:cf:constraintform:constraint} as a set of convex functions $\phi_i(\cdot)$:
\begin{equation}\label{eq:classifierconstraint}
    \phi_i(\x) \leq 0 \quad \forall\,i
\end{equation}
Substituting our proposed parametrization~\refeq{eq:cfparamaffinemapping} into~\refeq{eq:classifierconstraint} yields:
\begin{equation}\label{eq:classifierconstraint:new}
   \phi_i\left(f(\deltacf)\right) \leq 0 \quad \forall\,i 
\end{equation}
If our realization of the action mapping function~\refeq{eq:actionfunction} is convex - our realization for feature dependencies and plausibility are both affine and thus convex -, the original constraints~\refeq{eq:classifierconstraint} are convex and the concatenation of two convex functions is convex~\cite{Boyd2004}, the new constraints~\refeq{eq:classifierconstraint:new} are also convex.
Convexity of the objective~\refeq{eq:cfactionable:constraintform:objective} depends on the chosen regularization $\regularization(\cdot)$ - it is clear that convexity is given in case of the weighted Manhattan or L2 norm.

Our proposed modeling for actionable \& plausible counterfactuals therefore nicely fits into our previously proposed convex modeling framework for computing counterfactual explanations - i.e. we can use convex optimization for efficiently computing counterfactual explanations for many different standard ML models.

\section{Experiments}\label{sec:experiments}
We empirically verify our proposed modeling for actionable \& plausible counterfactuals by comparing them to unconstrained counterfactual and check if and how they differ. The Python implementation of the experiments is available on GitHub\footnote{\url{https://github.com/andreArtelt/ActionableCounterfactualsConvexProgramming}} whereby we use MOSEK\footnote{We gratefully acknowledge an academic license provided by MOSEK ApS.} as a solver for all mathematical programs.

\subsection{Feature dependencies}
\paragraph*{Data sets}
We use the following three standard benchmark data sets: The ``Iris Plants Data Set''~\cite{irisdata}, the ``Breast Cancer Wisconsin (Diagnostic) Data Set''~\cite{breastcancer} and the ``Wine data set''~\cite{winedata}.

\paragraph*{Models}
We use a softmax regression classifier and a general learning vector quantization  (GLVQ) classifier with $3$ prototypes per class.

\paragraph*{Setup}
We perform the following experiment over a $3$-fold cross validation:
In order to improve numerical stability, the training and test samples are standardized before the classifier is fitted and the correlation matrix is estimated from the training data - we use~\refeq{eq:sparsecovmatestimator} and $\alpha=0.8$ for estimating a sparse covariance matrix.
For all correctly classified samples in the test set, we compute a normal closest counterfactual~\refeq{eq:cf:constraintform} and a closest actionable counterfactual (under~\refeq{eq:cfparamaffinemapping}) - in both cases we use L1 norm as a regularization $\regularization(\cdot)$ and use a random target label.
For each pair of counterfactuals we compute their Euclidean distance and the number of overlapping features - we consider a feature to be overlapping if it is in both cases equal to zero or not equal to zero.

\begin{figure}[tb]
  \caption{Box plots of the number of overlapping features of closest counterfactuals and closest actionable counterfactuals.}
  \label{fig:expresults:featureoverlap}
  
  \centering
  
  \textbf{Softmax regression}
  
  \begin{minipage}[b]{0.3\textwidth}
    \includegraphics[width=\textwidth]{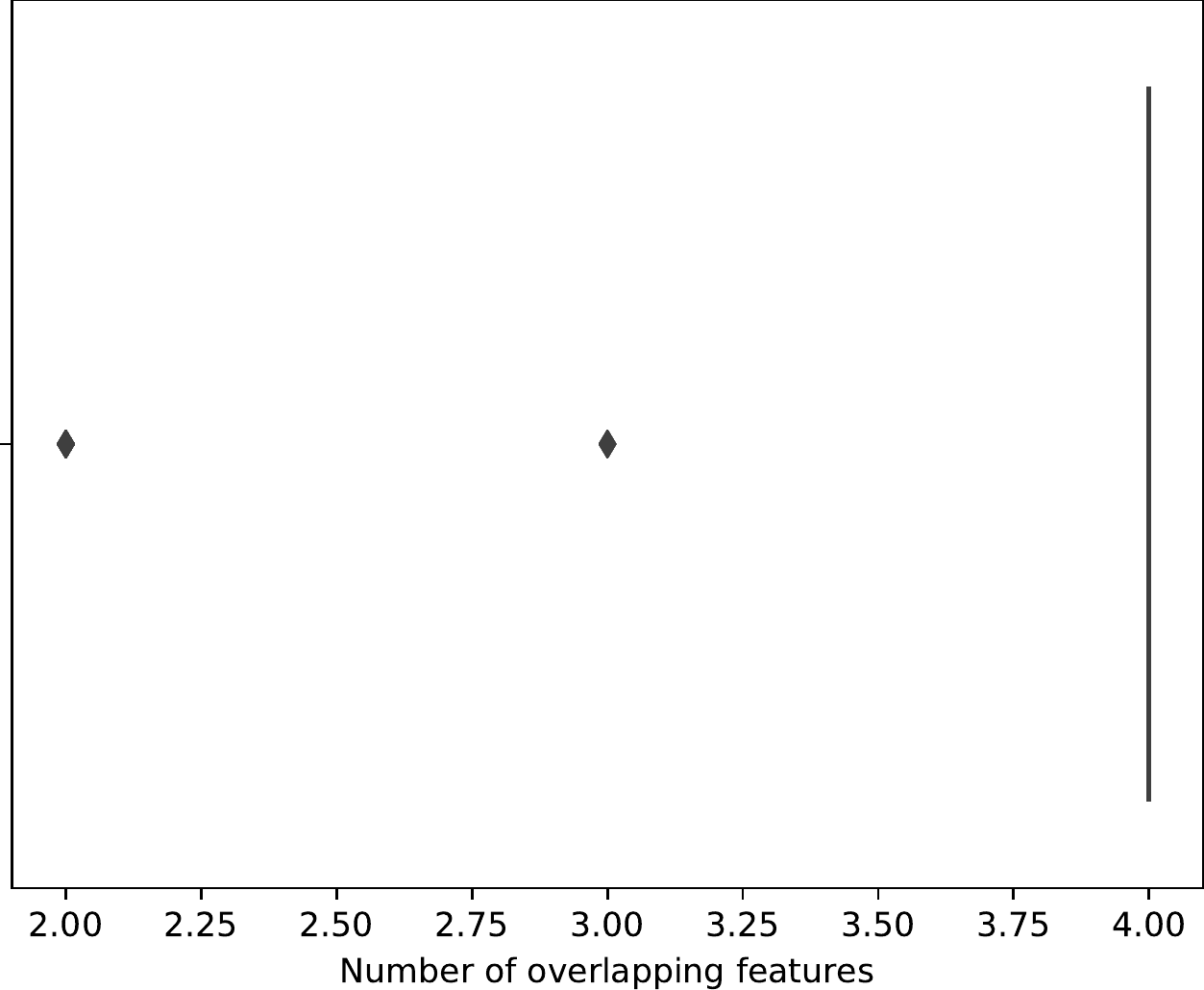}  
    \caption*{Iris} 
   \end{minipage}
  \hfill
  \begin{minipage}[b]{0.3\textwidth}
    \includegraphics[width=\textwidth]{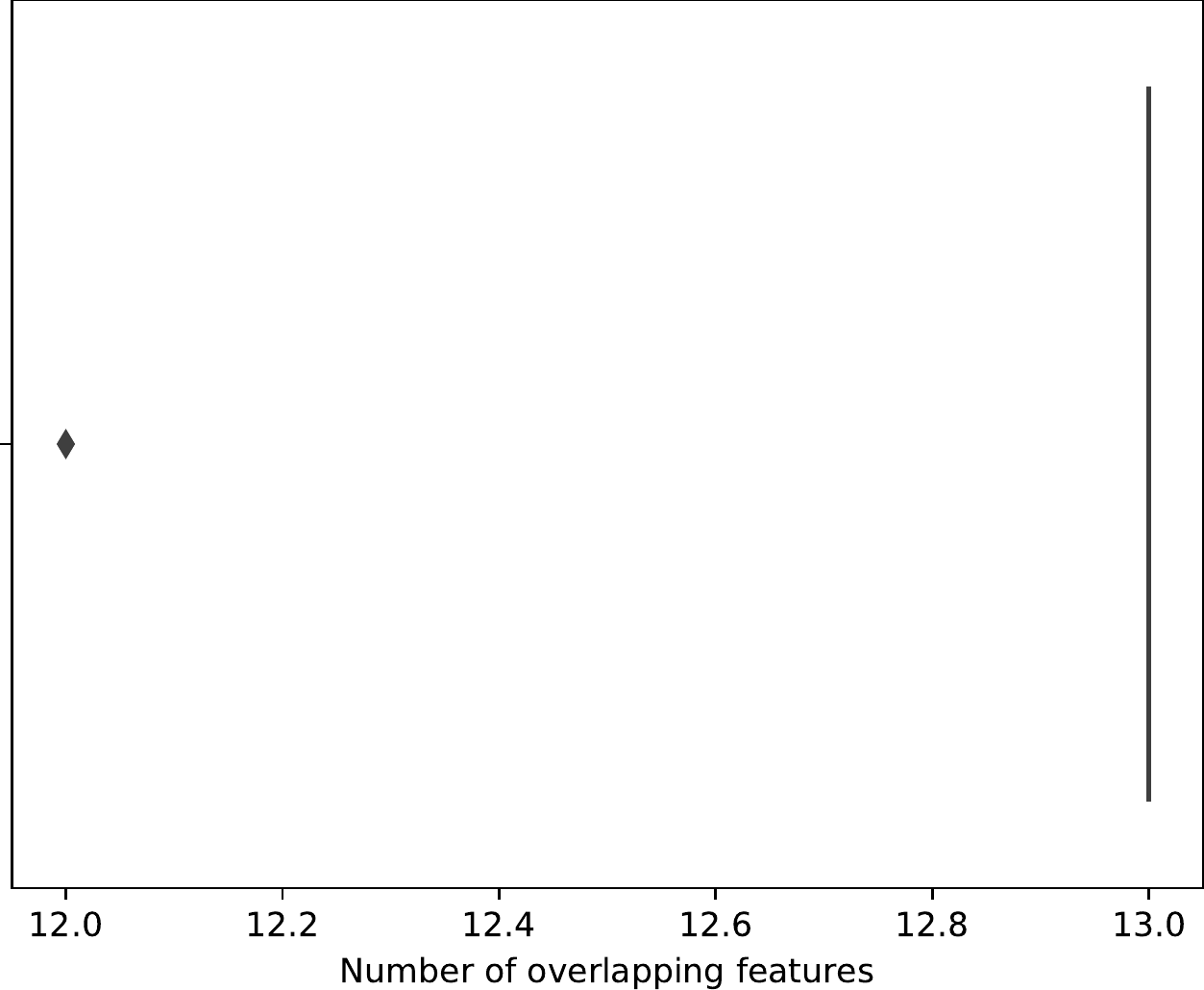}
    \caption*{Wine} 
  \end{minipage}
  \hfill
  \begin{minipage}[b]{0.3\textwidth}
    \includegraphics[width=\textwidth]{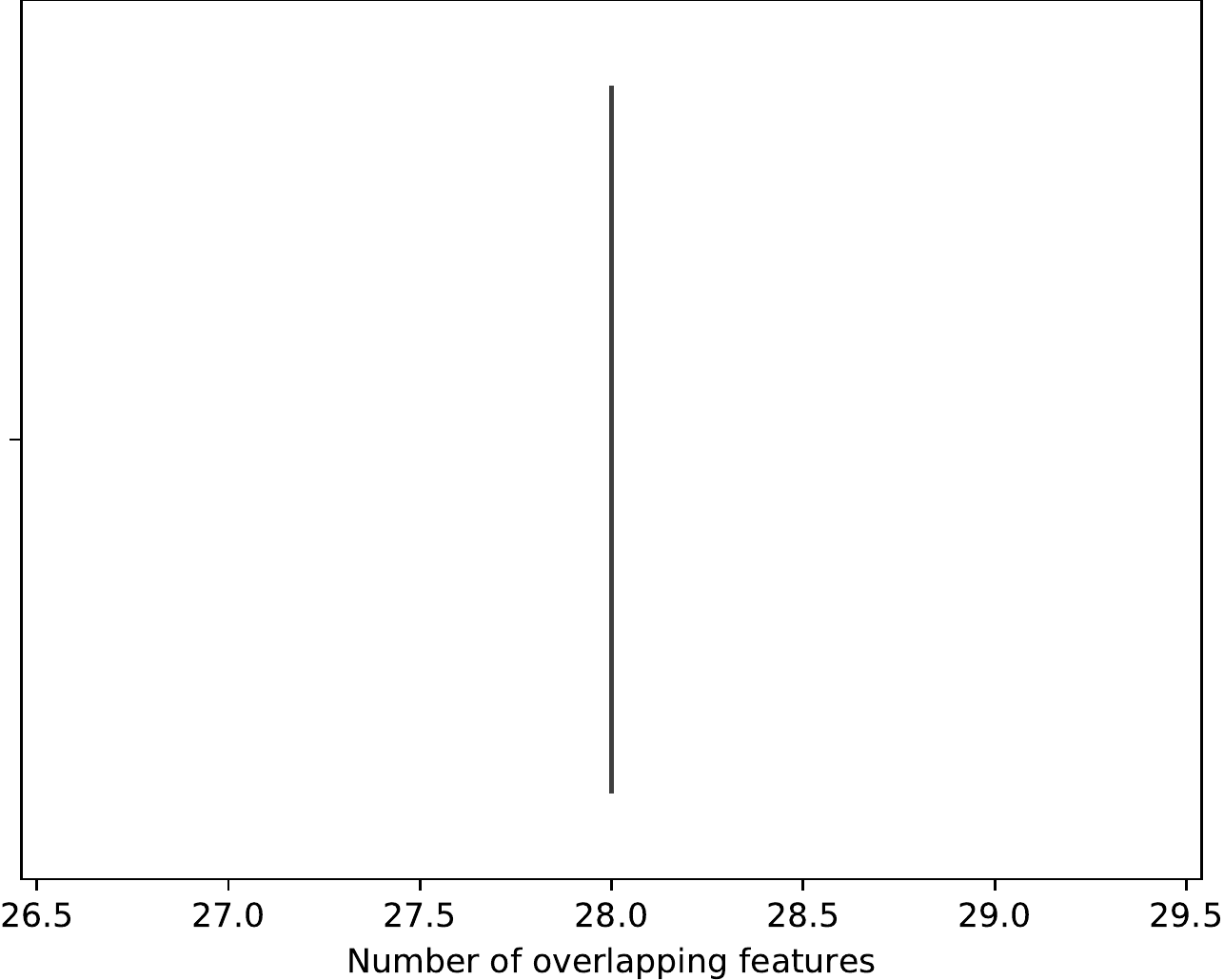}
    \caption*{Breast cancer} 
  \end{minipage}
 
  \textbf{GLVQ}
  \vfill
  \begin{minipage}[b]{0.3\textwidth}
    \includegraphics[width=\textwidth]{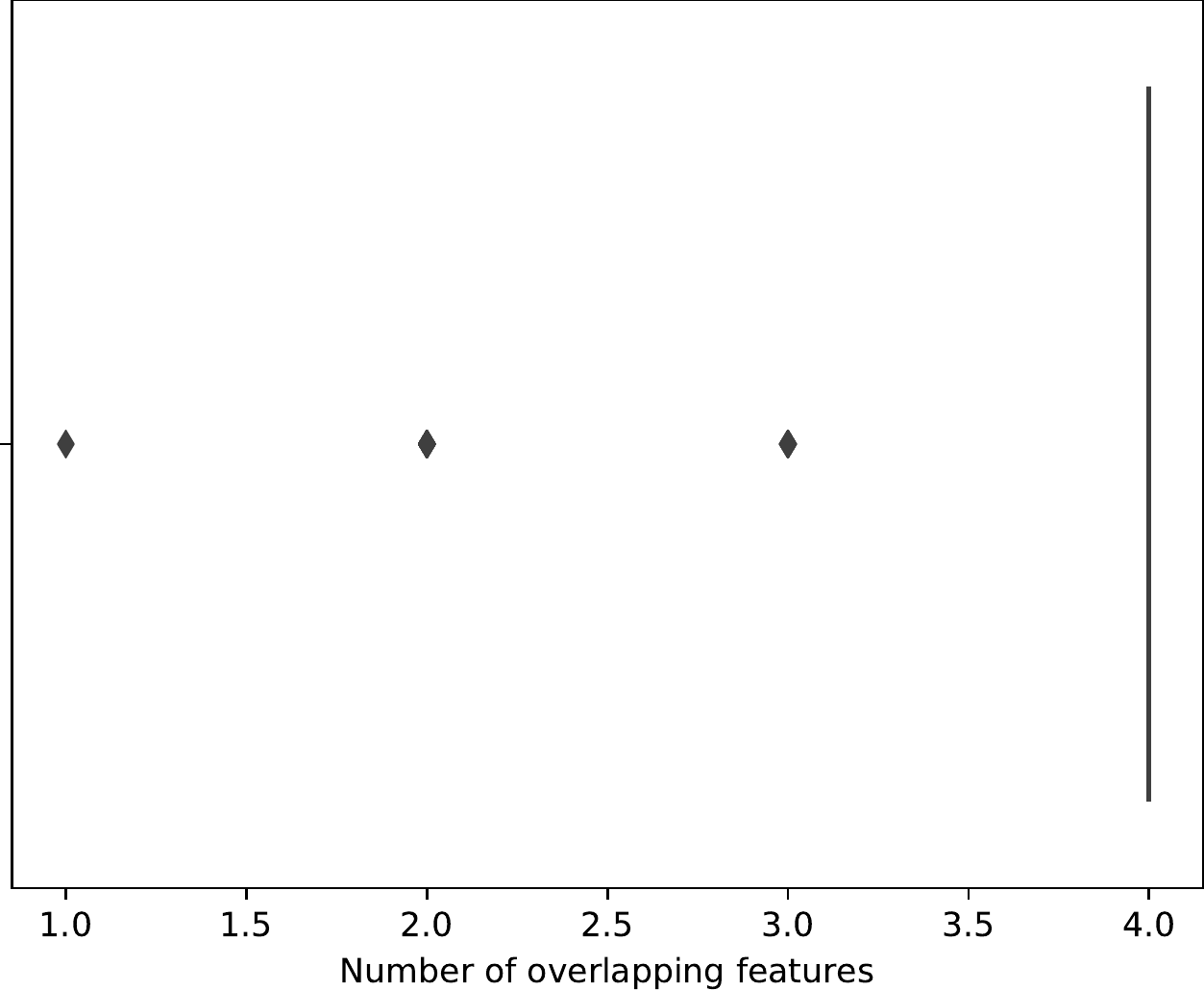}  
    \caption*{Iris} 
   \end{minipage}
  \hfill
  \begin{minipage}[b]{0.3\textwidth}
    \includegraphics[width=\textwidth]{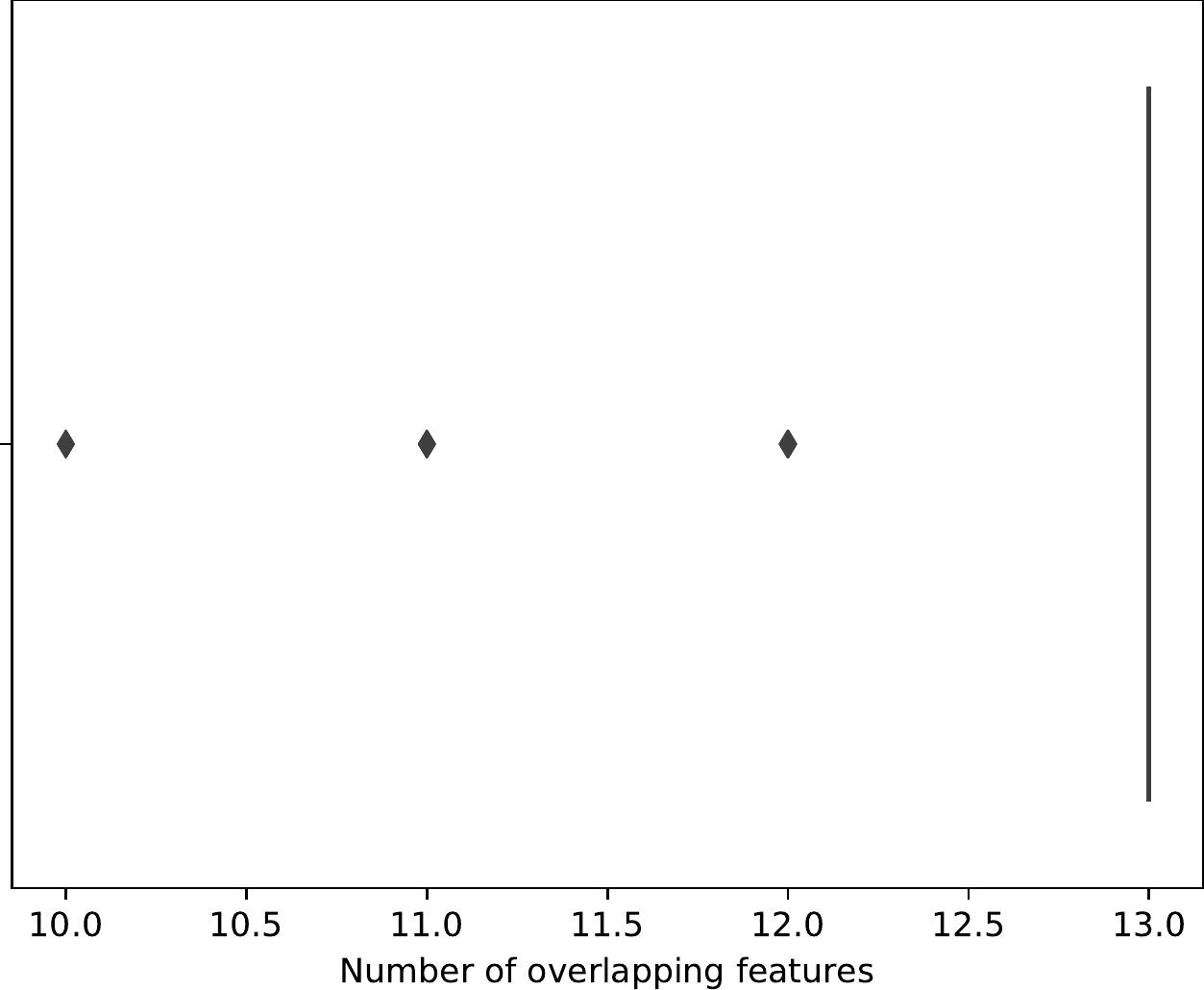}
    \caption*{Wine} 
  \end{minipage}
  \hfill
  \begin{minipage}[b]{0.3\textwidth}
    \includegraphics[width=\textwidth]{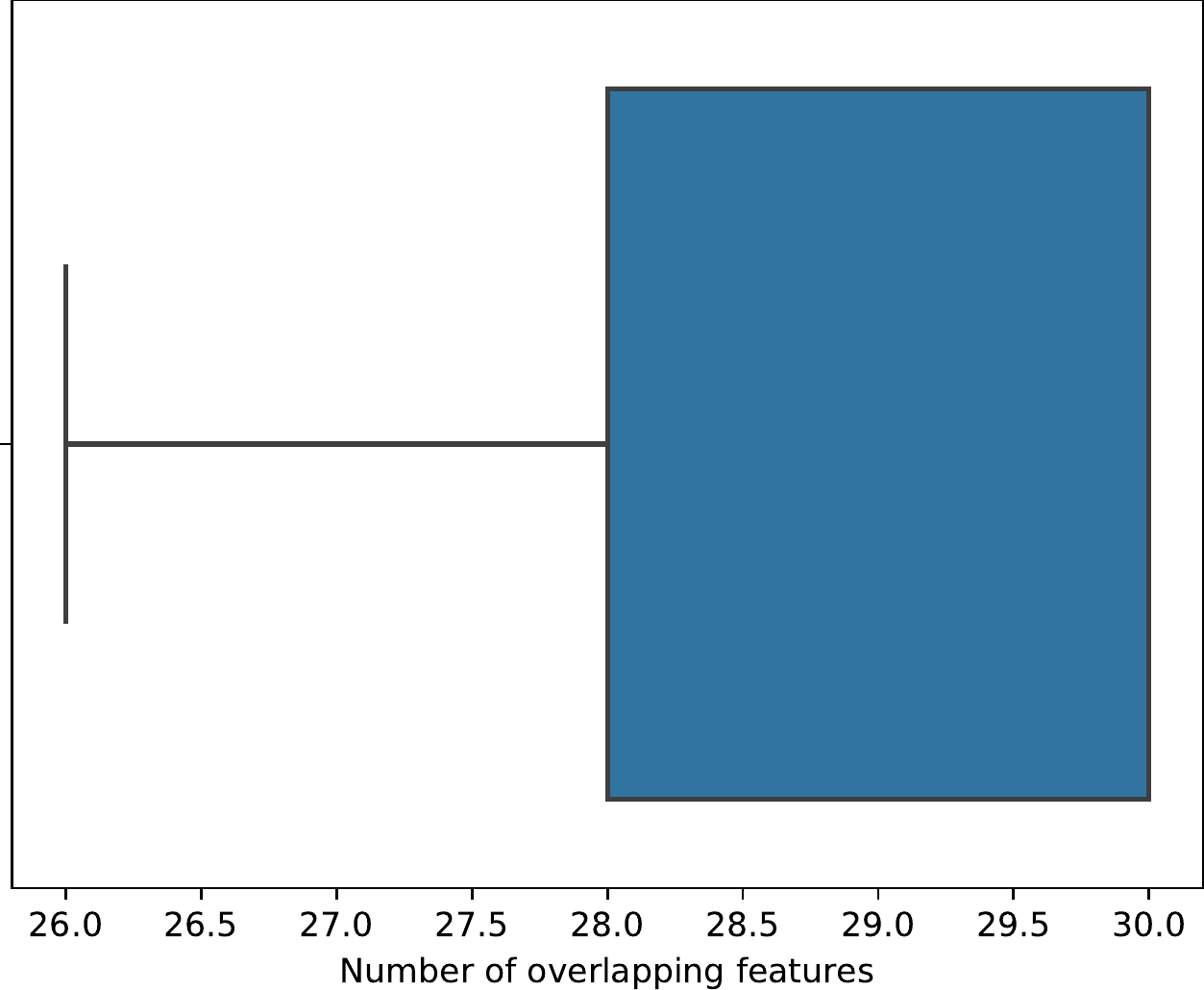}
    \caption*{Breast cancer} 
  \end{minipage}
\end{figure}

\paragraph*{Results}
The median Euclidean distances between the closest counterfactual and the closest actionable counterfactual is shown in Table~\ref{table:expresults:dist}.
\begin{table}[t]
\centering
\footnotesize
\begin{tabular}{|c||c||c||c|}
 \hline
 \textit{Data set} & Iris & Wine & Breast cancer  \\
 \hline
 Softmax regression & 0.74 & 0.1 & 11.12 \\
 GLVQ  & 0.63 & 0.62 & 11.52 \\
 \hline
\end{tabular}
\caption{Median \emph{Euclidean distance} between the closest counterfactual and the closest actionable counterfactual}
\label{table:expresults:dist}
\end{table}
On average, we observe in all cases significant differences between the closest counterfactual and the closest actionable counterfactional. This suggests that there might be substantial differences in both types of counterfactuals. However, the Euclidean distance itself does not tell anything about the chosen features - i.e. are the same or different features changed. Therefore, we plot the the number of overlapping features in Fig.~\ref{fig:expresults:featureoverlap} - recall that we consider a feature to be overlapping if it is in both cases equal to zero or not equal to zero (i.e. we count the number of same changed features).
We observe that in most cases the same features are changed - although there are always a few outliers. Only in case of the breast cancer data set we observe a significant difference in the choose features - on average there is a difference of two features. This shows that, depending on the data set, the covariance matrix and the model, it can actually happen that a closest actionable counterfactual uses different features than a closest counterfactual explanations.

\paragraph{Example}
For an illustrative (but real) example, consider the following correlation matrix that has been estimated from the Iris data set:
\begin{equation}\label{eq:example:corrmat}
    \begin{pmatrix}
        1. & 0. & 0.07913005 & 0.04025559 \\
        0 & 1. & 0. & 0. \\
        0.07913005 & 0. & 1. & 0.16517646 \\
        0.04025559 & 0. & 0.16517646 & 1.
    \end{pmatrix}
\end{equation}
When computing the closest counterfactual explanation (under a softmax regression) of a specific sample $\xorig$, we find that:
\begin{equation}
    \xcf = (-1.53442, 0., 0., 0. )^\top
\end{equation}
However, when considering the correlation matrix~\refeq{eq:example:corrmat}, the closest actionable counterfactual at the same sample $\xorig$ turns out to be:
\begin{equation}
    \xcf = (0., 0., 0., -1.36813)^\top
\end{equation}
Because of the correlations in~\refeq{eq:example:corrmat}, the selected feature to be changed is different for closest counterfactual vs. closest actionable counterfactual.

\subsection{Plausibility}
\begin{figure}[tb]
  \caption{Closest vs. plausible counterfactual explanations. First row shows the original samples, the closest and plausible counterfactuals are shown in the second and third row. The original label or target label is given below each image.}
  \label{fig:expresults:plausibility}
  
  \centering
  
  \textbf{Original samples}
  
  \begin{minipage}[b]{0.24\textwidth}
    \includegraphics[width=\textwidth]{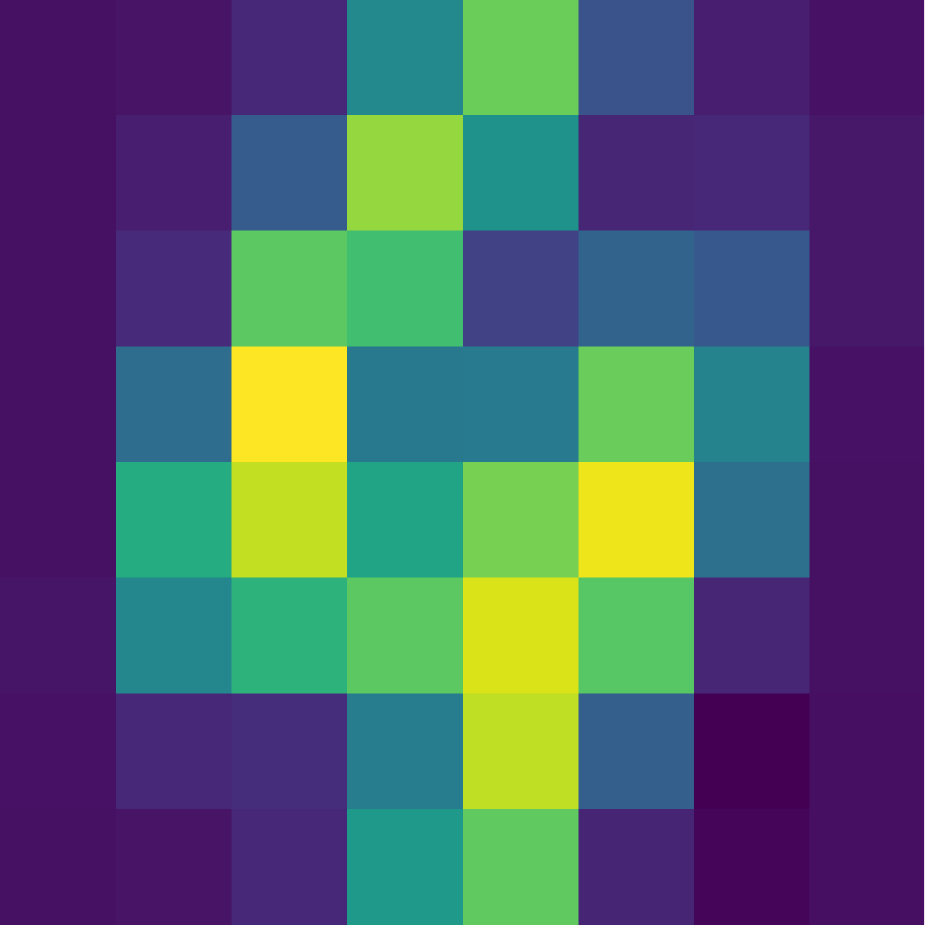}  
    \caption*{Label: 4} 
   \end{minipage}
  \hfill
  \begin{minipage}[b]{0.24\textwidth}
    \includegraphics[width=\textwidth]{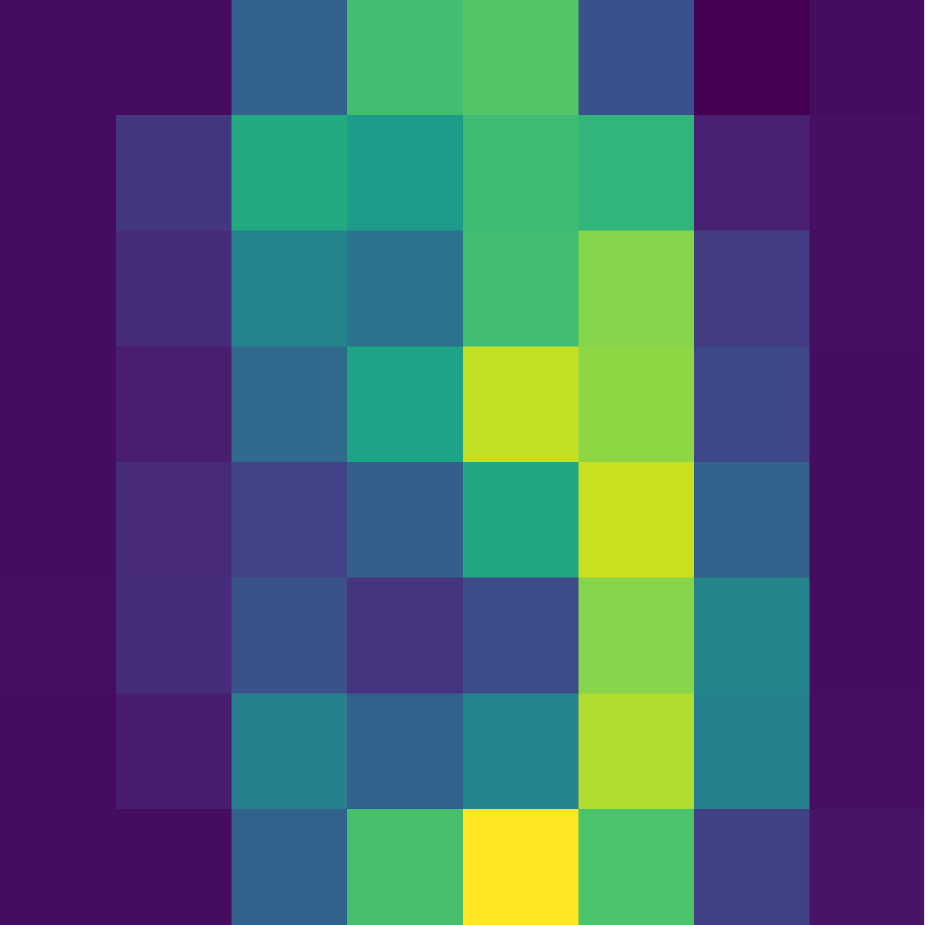}
    \caption*{Label: 3} 
  \end{minipage}
  \hfill
  \begin{minipage}[b]{0.24\textwidth}
    \includegraphics[width=\textwidth]{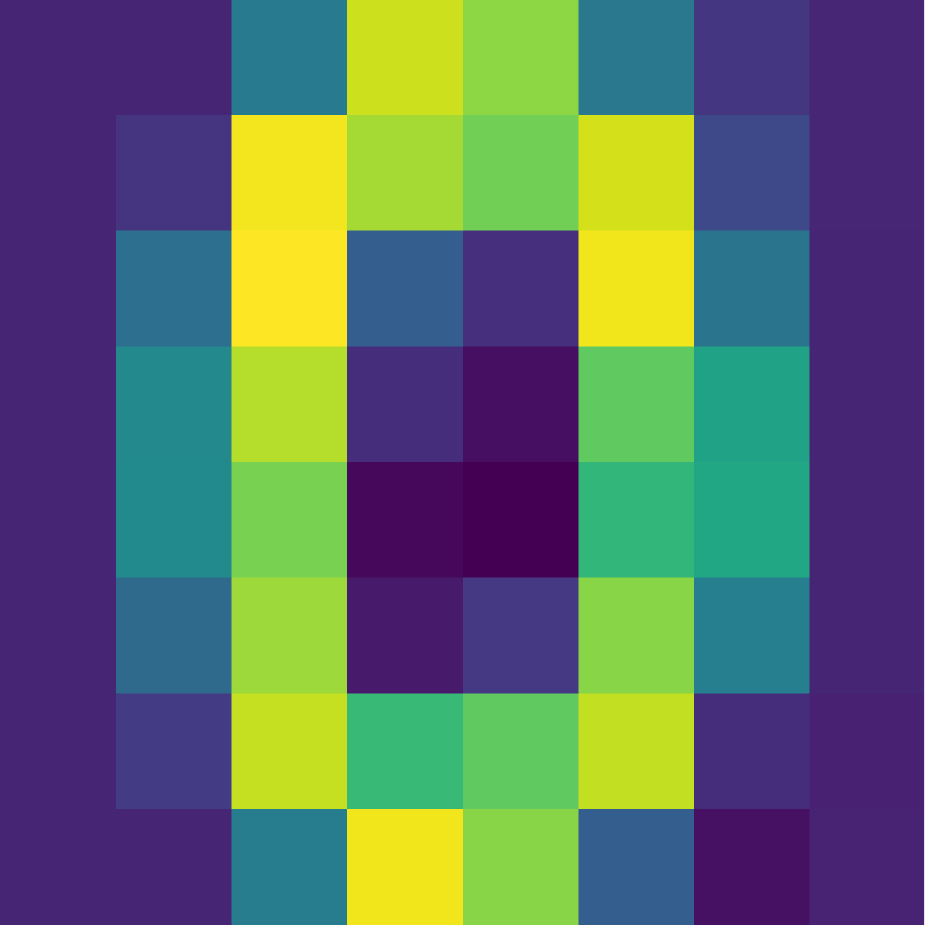}
    \caption*{Label: 0} 
  \end{minipage}
  \hfill
  \begin{minipage}[b]{0.24\textwidth}
    \includegraphics[width=\textwidth]{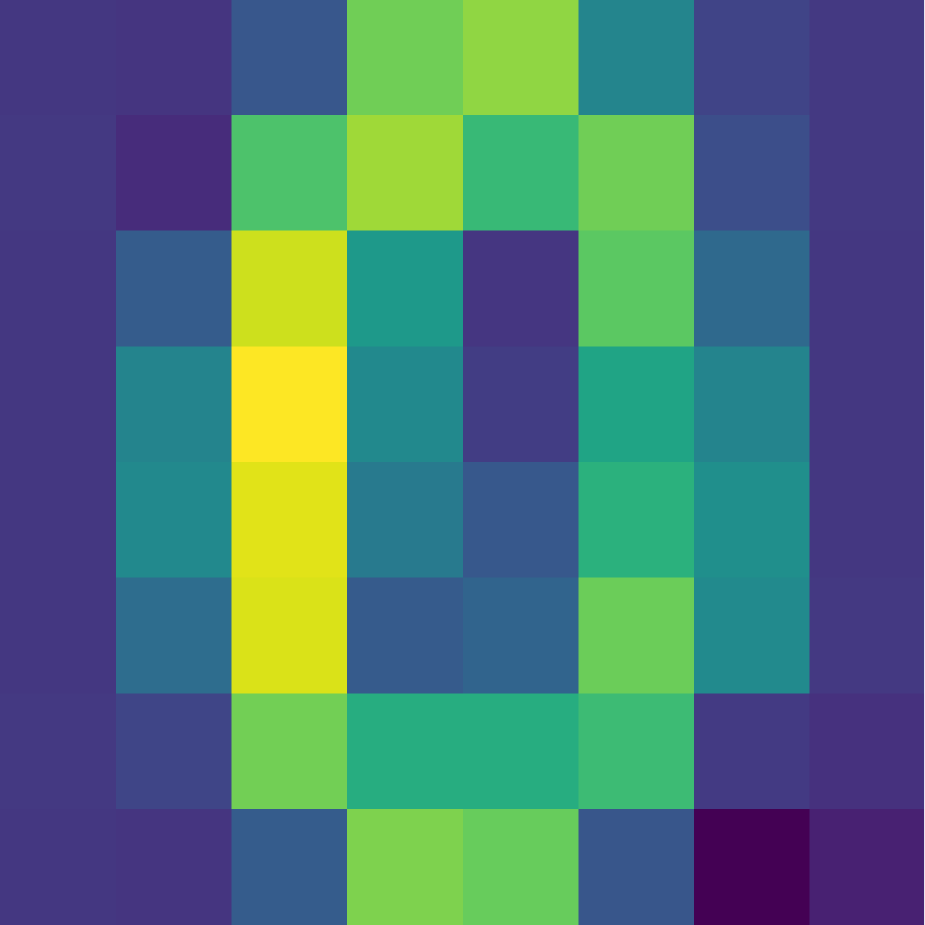}
    \caption*{Label: 0} 
  \end{minipage}
 
  \rule[2ex]{12.2cm}{2.0pt}  
 
  \textbf{Closest counterfactuals}
  \vfill
  \begin{minipage}[b]{0.24\textwidth}
    \includegraphics[width=\textwidth]{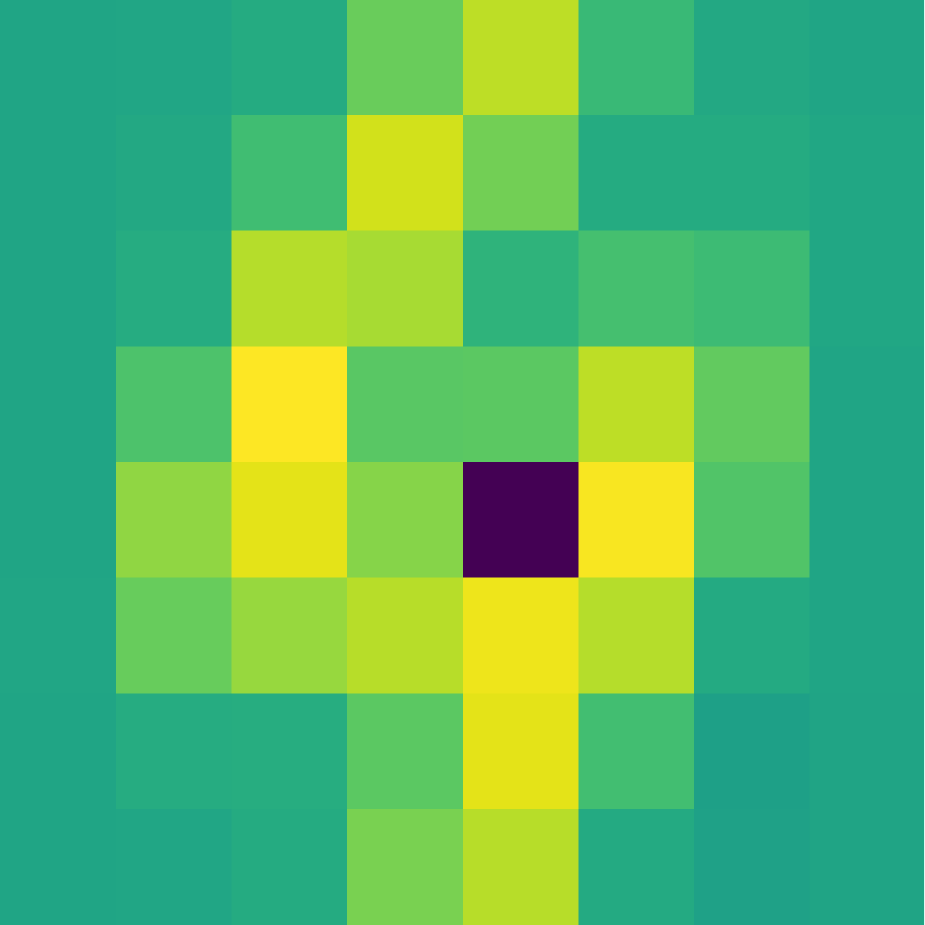}  
    \caption*{Label: 0} 
   \end{minipage}
  \hfill
  \begin{minipage}[b]{0.24\textwidth}
    \includegraphics[width=\textwidth]{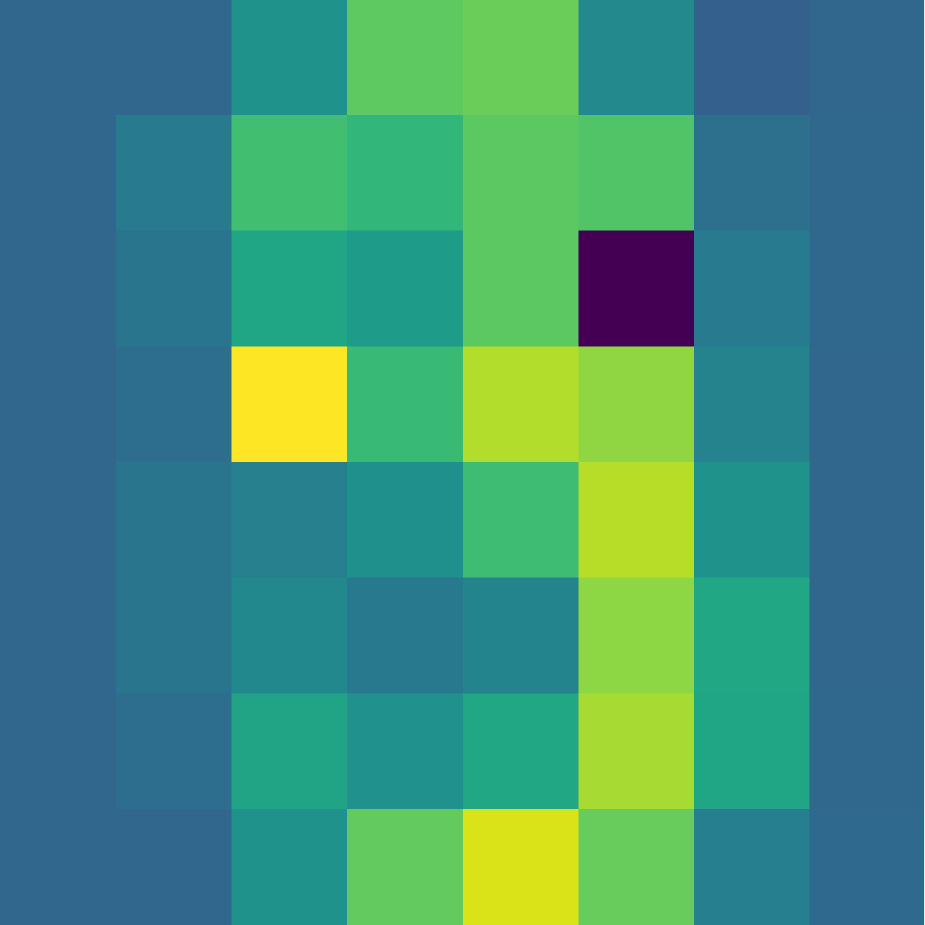}
    \caption*{Label: 5}  
  \end{minipage}
  \hfill
  \begin{minipage}[b]{0.24\textwidth}
    \includegraphics[width=\textwidth]{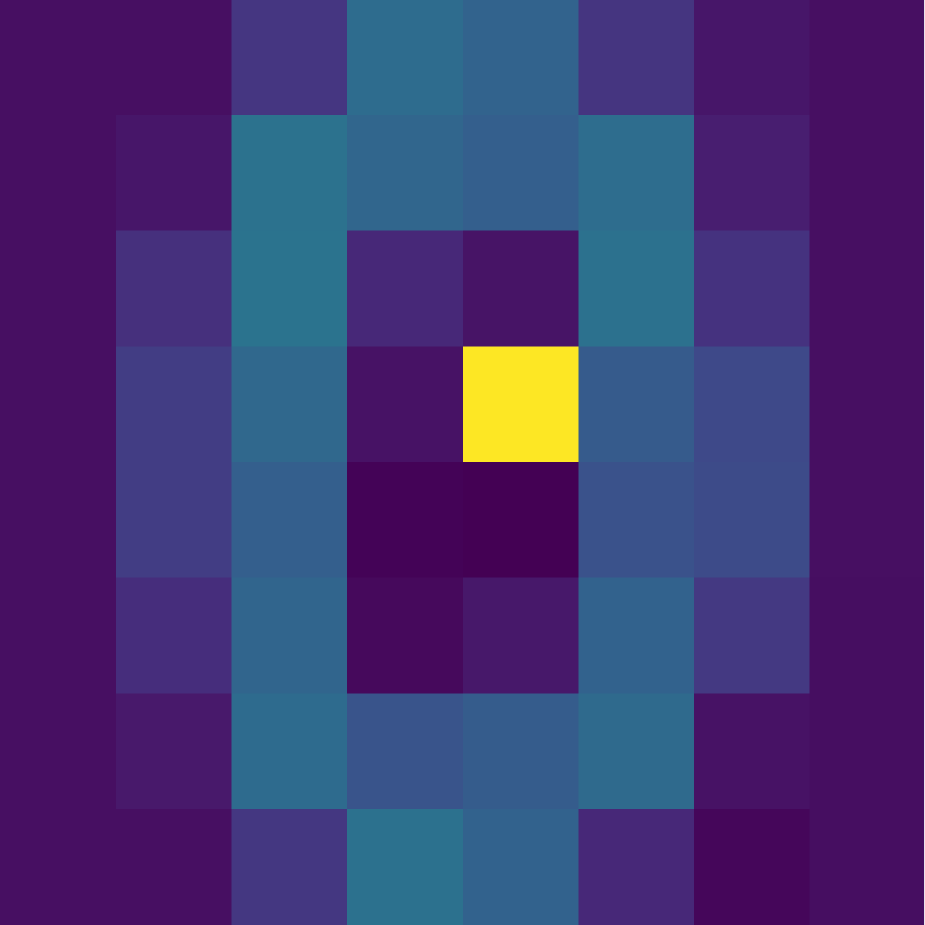}
    \caption*{Label: 9} 
  \end{minipage}
  \hfill
  \begin{minipage}[b]{0.24\textwidth}
    \includegraphics[width=\textwidth]{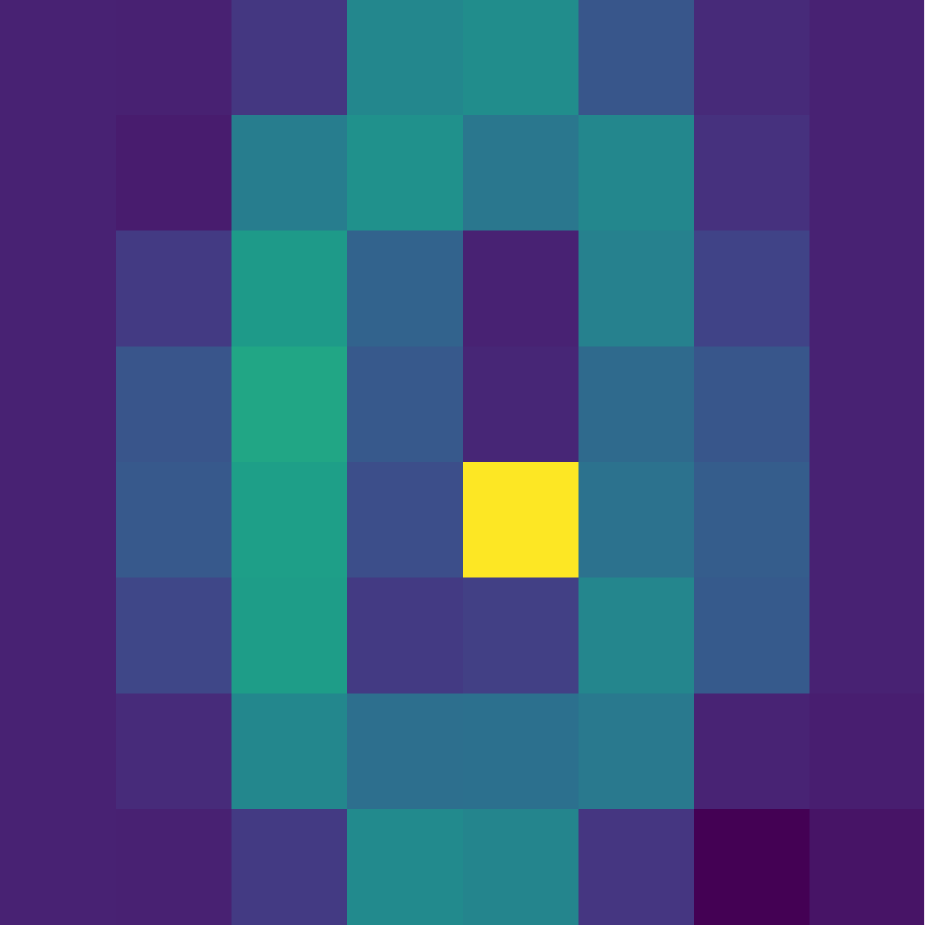}
    \caption*{Label: 6} 
  \end{minipage}
  
  \textbf{Plausible counterfactuals}
  \vfill
  \begin{minipage}[b]{0.24\textwidth}
    \includegraphics[width=\textwidth]{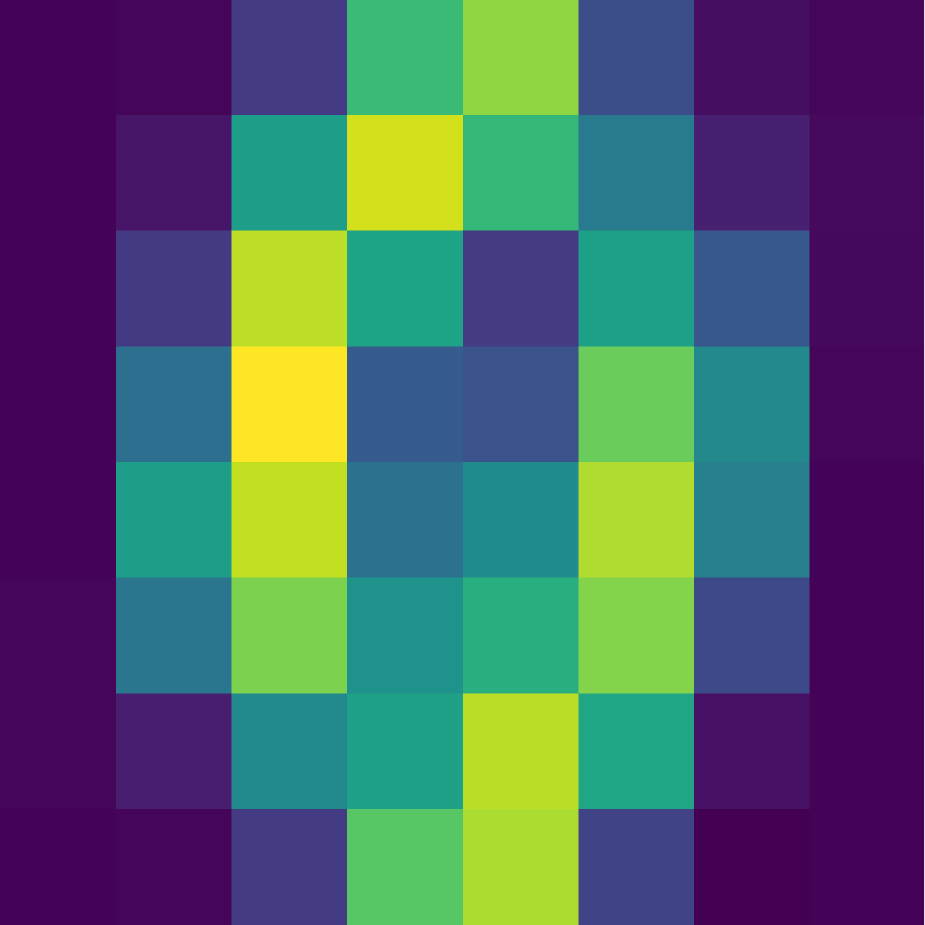}  
    \caption*{Label: 0} 
   \end{minipage}
  \hfill
  \begin{minipage}[b]{0.24\textwidth}
    \includegraphics[width=\textwidth]{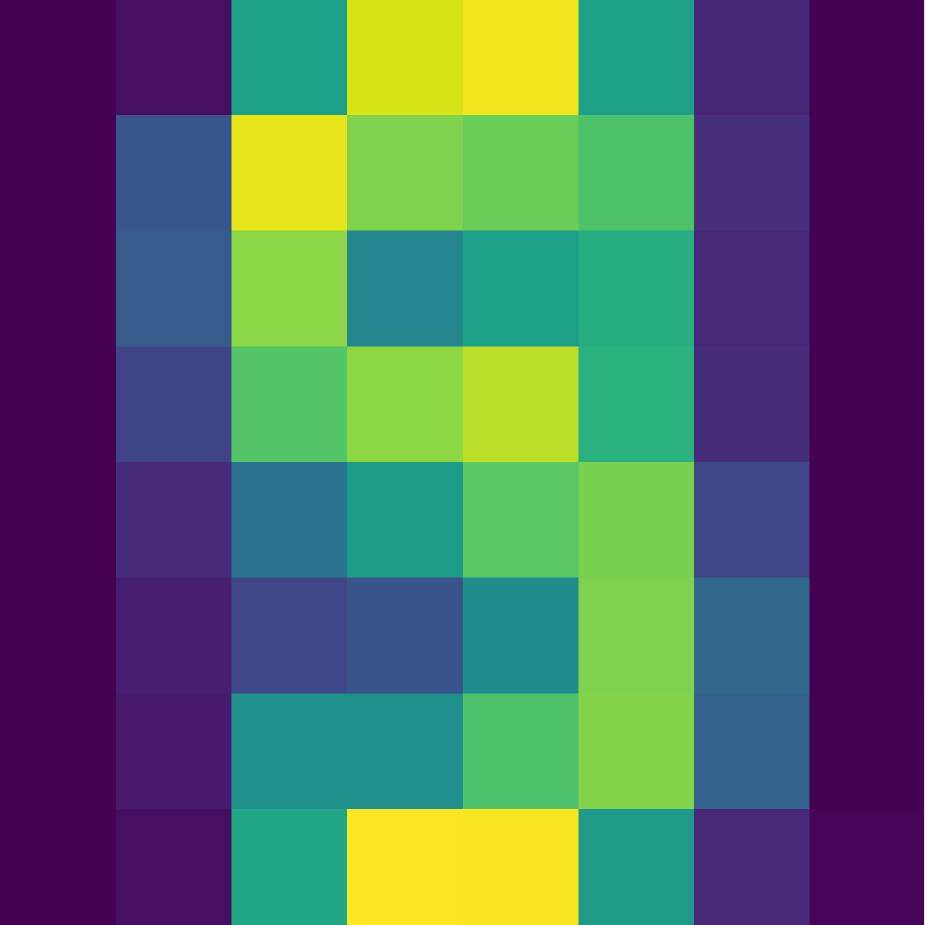}
    \caption*{Label: 5} 
  \end{minipage}
  \hfill
  \begin{minipage}[b]{0.24\textwidth}
    \includegraphics[width=\textwidth]{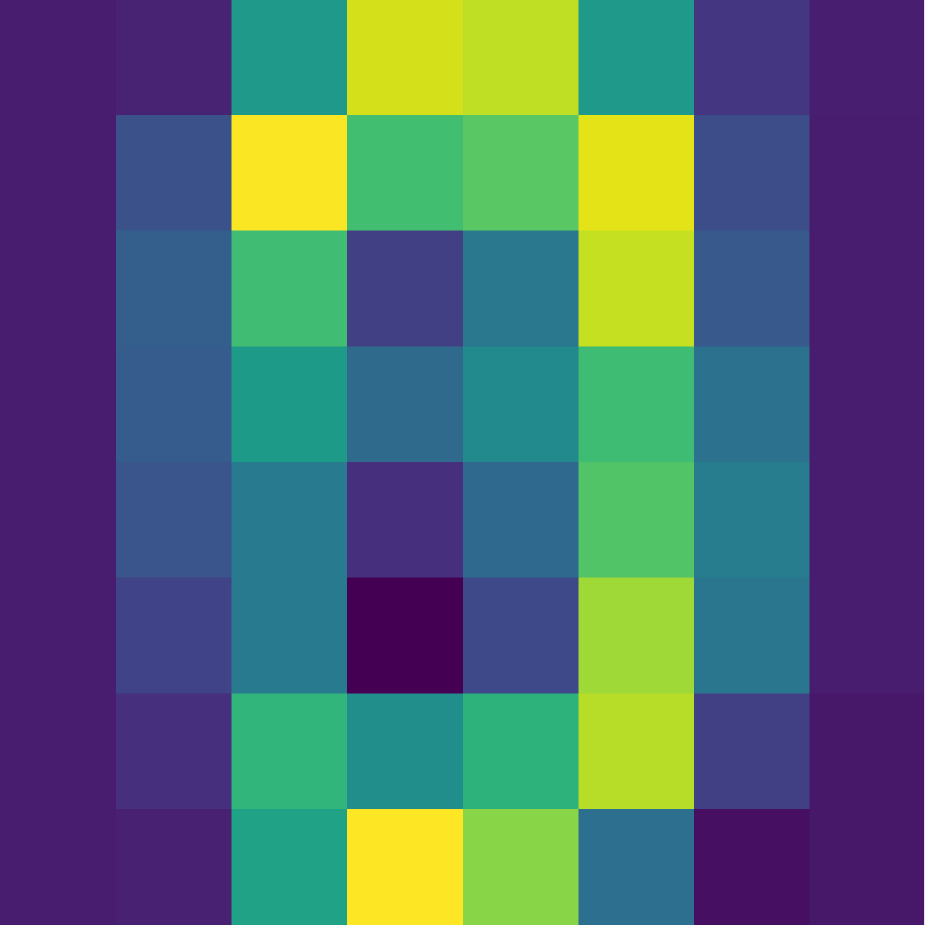}
    \caption*{Label: 9} 
  \end{minipage}
  \hfill
  \begin{minipage}[b]{0.24\textwidth}
    \includegraphics[width=\textwidth]{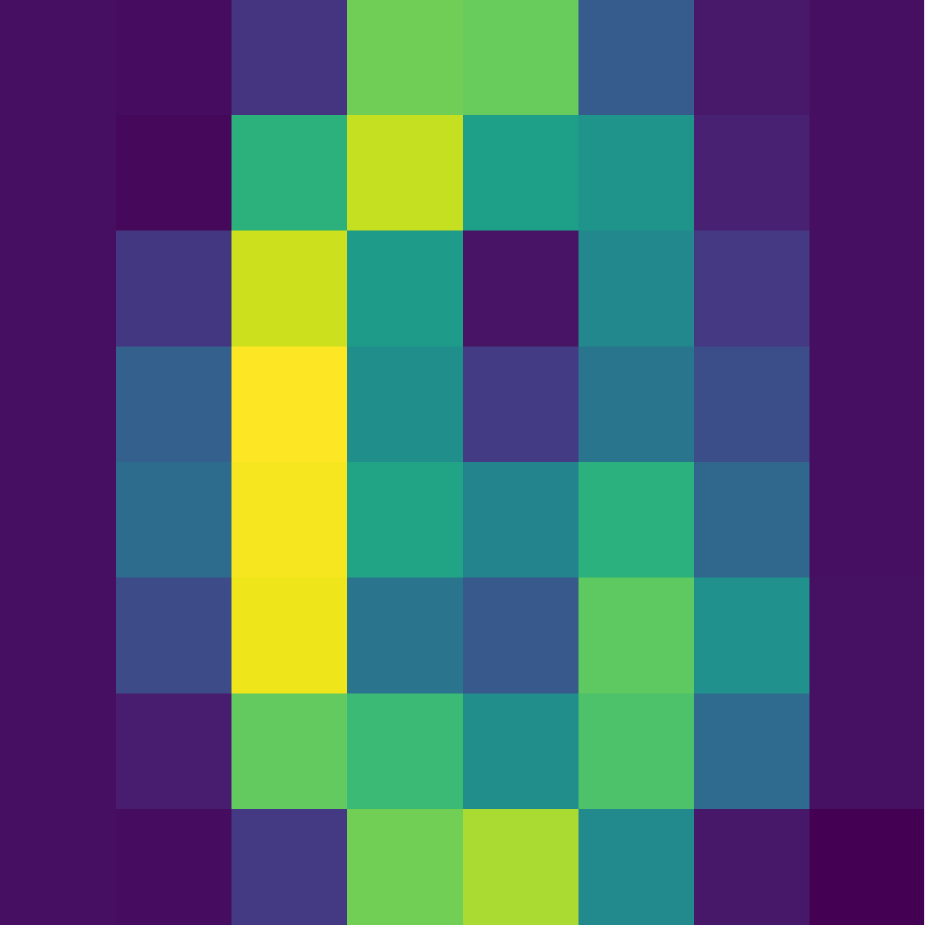}
    \caption*{Label: 6} 
  \end{minipage}
\end{figure}

In order to evaluate plausibility of the counterfactuals from section~\ref{sec:modeling:plausibility}, we compute closest counterfactuals and plausible counterfactuals (according to the action function~\refeq{eq:actionfunction:plausibility}) of the Digits data set~\cite{ocr} under a softmax regression model. We use sparse dictionary learning to learn $10$ prototypes/primitives that are used to encode all samples.
We show a few examples of original, closest and plausible counterfactuals in Fig.~\ref{fig:expresults:plausibility}.
We observe that most cases the closest counterfactual looks like an adversarial whereas in case of the plausible counterfactual the target class can be often clearly recognized. However, note that we are looking at a few samples only and a ``true'' evaluation would require an extensive user study since perception of plausibility is difficult to formalize.

\section{Conclusion}\label{sec:conclusion}
In this work we extended our convex programming framework for computing counterfactual explanations by a mechanism where we optimize over the actions rather than directly the final counterfactual explanations. By this we were able to consider linear feature dependencies - i.e. getting rid of the assumption that all features are independent -, as well as ensuring plausibility of the final counterfactual by constructing it  as a linear combination of given primitives/prototypes. Besides working out the modeling details, we also empirically evaluated our modeling on different data sets, where we observed significant differences in the counterfactuals considering estimated correlations versus counterfactuals assuming feature independence.

Since our formalization is limited to linear dependencies we plan to extend our modeling for non-linear dependencies as well as to consider more ML models like deep neural networks. We also would like to study the difference of closest and closest actionable counterfactuals from a psychological point of view - i.e. how are they perceived and in particular whether closest actionable counterfactuals are considered to be ``better'' or not.


\begin{footnotesize}


\bibliographystyle{unsrt}
\bibliography{bibliography}

\begin{thebibliography}{10}

\bibitem{PredictivePolicing}
Julia Angwin, Jeff Larson, Surya Mattu, and Lauren Kirchner.
\newblock Machine bias - there's software used across the country to predict
  future criminals. and it's biased against blacks.
\newblock 2016.

\bibitem{CreditRiskML}
Amir~E. Khandani, Adlar~J. Kim, and Andrew Lo.
\newblock Consumer credit-risk models via machine-learning algorithms.
\newblock {\em Journal of Banking \& Finance}, 34(11), 2010.

\bibitem{CreditScoresUnfair}
Kaveh Waddell.
\newblock How algorithms can bring down minorities' credit scores.
\newblock {\em The Atlantic}, 2016.

\bibitem{gdpr}
European parliament and council.
\newblock Regulation (eu) 2016/679 of the european parliament and of the
  council of 27 april 2016 on the protection of natural persons with regard to
  the processing of personal data and on the free movement of such data, and
  repealing directive 95/46/ec (general data protection regulation).
\newblock \url{https://eur-lex.europa.eu/eli/reg/2016/679/oj}, 2016.

\bibitem{featureinteraction}
Brandon~M. Greenwell, Bradley~C. Boehmke, and Andrew~J. McCarthy.
\newblock A simple and effective model-based variable importance measure.
\newblock {\em CoRR}, abs/1805.04755, 2018.

\bibitem{featureimportance}
Aaron {Fisher}, Cynthia {Rudin}, and Francesca {Dominici}.
\newblock {All Models are Wrong but many are Useful: Variable Importance for
  Black-Box, Proprietary, or Misspecified Prediction Models, using Model Class
  Reliance}.
\newblock {\em arXiv e-prints}, page arXiv:1801.01489, Jan 2018.

\bibitem{casebasedreasoning}
A.~Aamodt and E.~Plaza.
\newblock Case-based reasoning: Foundational issues, methodological variations,
  and systemapproaches.
\newblock {\em AI communications}, 1994.

\bibitem{influentialinstances}
Pang~Wei Koh and Percy Liang.
\newblock Understanding black-box predictions via influence functions.
\newblock In {\em Proceedings of the 34th International Conference on Machine
  Learning, {ICML} 2017, Sydney, NSW, Australia, 6-11 August 2017}, pages
  1885--1894, 2017.

\bibitem{prototypescriticism}
Been Kim, Oluwasanmi Koyejo, and Rajiv Khanna.
\newblock Examples are not enough, learn to criticize! criticism for
  interpretability.
\newblock In {\em Advances in Neural Information Processing Systems 29: Annual
  Conference on Neural Information Processing Systems 2016, December 5-10,
  2016, Barcelona, Spain}, pages 2280--2288, 2016.

\bibitem{counterfactualwachter}
Sandra Wachter, Brent~D. Mittelstadt, and Chris Russell.
\newblock Counterfactual explanations without opening the black box: Automated
  decisions and the {GDPR}.
\newblock {\em CoRR}, abs/1711.00399, 2017.

\bibitem{molnar2019}
Christoph Molnar.
\newblock {\em Interpretable Machine Learning}.
\newblock 2019.
\newblock \url{https://christophm.github.io/interpretable-ml-book/}.

\bibitem{ijcai2019-876}
Ruth M.~J. Byrne.
\newblock Counterfactuals in explainable artificial intelligence (xai):
  Evidence from human reasoning.
\newblock In {\em Proceedings of the Twenty-Eighth International Joint
  Conference on Artificial Intelligence, {IJCAI-19}}, pages 6276--6282.
  International Joint Conferences on Artificial Intelligence Organization, 7
  2019.

\bibitem{counterfactualcomputationsurvey}
Andr{\'{e}} Artelt and Barbara Hammer.
\newblock On the computation of counterfactual explanations - {A} survey.
\newblock {\em CoRR}, abs/1911.07749, 2019.

\bibitem{Boyd2004}
Stephen Boyd and Lieven Vandenberghe.
\newblock {\em Convex Optimization}.
\newblock Cambridge University Press, New York, NY, USA, 2004.

\bibitem{plausiblecounterfactualsartelt}
Andr\'e Artelt and Barbara Hammer.
\newblock Convex density constraints for computing plausible counterfactual
  explanations.
\newblock {\em 29th International Conference on Artificial Neural Networks
  (ICANN)}, 2020.

\bibitem{chou2021counterfactuals}
Yu-Liang Chou, Catarina Moreira, Peter Bruza, Chun Ouyang, and Joaquim Jorge.
\newblock Counterfactuals and causability in explainable artificial
  intelligence: Theory, algorithms, and applications, 2021.

\bibitem{face}
Rafael Poyiadzi, Kacper Sokol, Ra{\'{u}}l Santos{-}Rodriguez, Tijl~De Bie, and
  Peter~A. Flach.
\newblock {FACE:} feasible and actionable counterfactual explanations.
\newblock {\em CoRR}, abs/1909.09369, 2019.

\bibitem{galhotra2021explaining}
Sainyam Galhotra, Romila Pradhan, and Babak Salimi.
\newblock Explaining black-box algorithms using probabilistic contrastive
  counterfactuals, 2021.

\bibitem{DBLP:journals/corr/abs-1912-03277}
Divyat Mahajan, Chenhao Tan, and Amit Sharma.
\newblock Preserving causal constraints in counterfactual explanations for
  machine learning classifiers.
\newblock {\em CoRR}, abs/1912.03277, 2019.

\bibitem{DBLP:conf/fat/UstunSL19}
Berk Ustun, Alexander Spangher, and Yang Liu.
\newblock Actionable recourse in linear classification.
\newblock In danah boyd and Jamie~H. Morgenstern, editors, {\em Proceedings of
  the Conference on Fairness, Accountability, and Transparency, FAT* 2019,
  Atlanta, GA, USA, January 29-31, 2019}, pages 10--19. {ACM}, 2019.

\bibitem{DBLP:conf/ecai/FeghahatiSPT20}
Amir Feghahati, Christian~R. Shelton, Michael~J. Pazzani, and Kevin Tang.
\newblock Cdeepex: Contrastive deep explanations.
\newblock In Giuseppe~De Giacomo, Alejandro Catal{\'{a}}, Bistra Dilkina,
  Michela Milano, Sen{\'{e}}n Barro, Alberto Bugar{\'{\i}}n, and
  J{\'{e}}r{\^{o}}me Lang, editors, {\em {ECAI} 2020 - 24th European Conference
  on Artificial Intelligence, 29 August-8 September 2020, Santiago de
  Compostela, Spain, August 29 - September 8, 2020 - Including 10th Conference
  on Prestigious Applications of Artificial Intelligence {(PAIS} 2020)}, volume
  325 of {\em Frontiers in Artificial Intelligence and Applications}, pages
  1143--1151. {IOS} Press, 2020.

\bibitem{DBLP:journals/corr/abs-2103-10226}
Pau Rodr{\'{\i}}guez, Massimo Caccia, Alexandre Lacoste, Lee Zamparo, Issam~H.
  Laradji, Laurent Charlin, and David V{\'{a}}zquez.
\newblock Beyond trivial counterfactual explanations with diverse valuable
  explanations.
\newblock {\em CoRR}, abs/2103.10226, 2021.

\bibitem{DBLP:journals/corr/abs-2012-09301}
Rachana Balasubramanian, Samuel Sharpe, Brian Barr, Jason~D. Wittenbach, and
  C.~Bayan Bruss.
\newblock Latent-cf: {A} simple baseline for reverse counterfactual
  explanations.
\newblock {\em CoRR}, abs/2012.09301, 2020.

\bibitem{sparsecovmat}
Jerome Friedman, Trevor Hastie, and Robert Tibshirani.
\newblock {Sparse inverse covariance estimation with the graphical lasso}.
\newblock {\em Biostatistics}, 9(3):432--441, 12 2007.

\bibitem{irisdata}
Ronald~Aylmer Fisher.
\newblock The use of multiple measurements in taxonomic problems.
\newblock {\em Annual Eugenics}, 7 Part II:179--188, 1936.

\bibitem{breastcancer}
Olvi L.~Mangasarian William H.~Wolberg, W. Nick~Street.
\newblock Breast cancer wisconsin (diagnostic) data set.
\newblock
  \url{https://archive.ics.uci.edu/ml/datasets/Breast+Cancer+Wisconsin+(Diagnostic)},
  1995.

\bibitem{winedata}
D.~Coomans S.~Aeberhard and O.~de~Vel.
\newblock Comparison of classifiers in high dimensional settings.
\newblock {\em Tech. Rep. no. 92-02}, 1992.

\bibitem{ocr}
E.~Alpaydin and C.~Kaynak.
\newblock Optical recognition of handwritten digits data set.
\newblock
  \url{https://archive.ics.uci.edu/ml/datasets/Optical+Recognition+of+Handwritten+Digits},
  1998.

\end{thebibliography}

\end{footnotesize}


\end{document}